\documentclass{sig-alternate}
\newcommand{\eat}[1]{}
\usepackage{epsfig}
\usepackage{subfigure}
\usepackage{latexsym}
\usepackage{amssymb}
\usepackage{amsmath}
\usepackage{url}
\usepackage{colortbl}

\usepackage{float}
\usepackage{graphicx}
\usepackage[boxruled]{algorithm2e}
\usepackage{algorithmic}
\usepackage{amsmath}
\usepackage{pdfsync}
\usepackage{booktabs}
\usepackage{multirow}
\usepackage{textcomp}
\renewcommand{\footnotesize}{\tiny}

\begin{document}

\title{Temporal Embedding in Convolutional Neural Networks for Robust Learning of Abstract Snippets}

\author{
 $^\dag$Jiajun Liu \hspace{0.8cm} $^\dag$Kun Zhao \hspace{0.8cm}
$^\dag$Brano Kusy \hspace{0.8cm}
  $^*$Ji-rong Wen \hspace{0.8cm}
 $^\dag$Raja Jurdak
 \\
{$^\dag$AS Program, CSIRO, Pullenvale, Australia}\\
{ \{jiajun.liu, kun.zhao, brano.kusy, raja.jurdak\}@csiro.au}\\
{$^*$Renmin University of China, Beijing, China}\\
{jrwen@ruc.edu.cn}
%\hspace{-1.5cm}\affaddr{$^\dag$The University of Queensland \hspace{0.2cm} $^\#$Hong Kong University of \hspace{0.2cm} $^*$Ecole Polytechnique Federale de Lausanne}\\
%\hspace{-3cm}\affaddr{\{jiajun,huang,shenht\}@itee.uq.edu.au \hspace{0.4cm} Science and Technology  \hspace{1.5cm} }
}

\pagenumbering{arabic}
\setcounter{page}{1}

\maketitle

\begin{abstract}
The prediction of periodical time-series remains challenging due to various types of data distortions and misalignments. Here, we propose
 a novel  model called Temporal embedding-enhanced convolutional neural Network (TeNet) to learn repeatedly-occurring-yet-hidden structural elements in periodical  time-series, called abstract snippets,   for predicting future changes. 
Our model uses convolutional neural networks and embeds  a time-series with its potential neighbors in the temporal domain for aligning it to the dominant patterns in the dataset. The model is robust to distortions and misalignments in the temporal domain and demonstrates strong prediction power for periodical time-series.

We conduct extensive experiments and discover that the proposed model shows significant and consistent advantages over existing methods on a variety of data modalities ranging from human mobility to household power consumption records. 
Empirical results indicate that the model is robust to various factors such as number of samples, variance of data, numerical ranges of data etc. The experiments also verify that the intuition behind the model can be generalized to multiple data types and applications
and promises significant improvement in prediction performances across the datasets studied.
\end{abstract}

\section{Introduction}
\label{sec:intro}
The behaviors of many of the world's inhabitants are fundamentally bound by the
cycle of the sun and the moon which creates day and night. It is the
reason why across the days of an average person, there often exist
periodical patterns for their mobility or more generally, their behavior
\cite{schneider2013unravelling,schneider2013daily}. 
Utilizing such re-occurring patterns could drastically benefit various 
modern ubiquitous applications. For example, \eat{being able to accurately predict how much further a
person will be traveling for the rest of the day will give a location
tracking system a great advantage for planning its energy-aware sampling
strategy. Similarly, }the ability to predict a day's power consumption
of many individual houses at midday will be profoundly beneficial
for the smart grid to manage dynamically its power supply resources.   
While in the scenario of smart location tracking \cite{Jurdak13,jrsi}, with a replenish-able energy budget the system
 either aims to minimize the energy efficiency of location tracking,
 or attempts to maximize the tracking accuracy given a fixed energy
 budget. A crucial challenge involved in such a smart tracking system is to
 estimate at any time  of day how much further the moving entities will move for the
 remainder of the day. Ideally, with a greater estimated value of the total travel distance, 
 the system will employ a more conservative sampling strategy (lower sampling frequencies) to 
 cover as much as possible of the whole trip using the restricted energy budget, 
 whereas a more aggressive strategy (higher sampling frequencies) will be favored on the
 presence of a smaller estimated total travel distance, so that better tracking precision will be achieved. 
 Clearly the estimation of the entity's daily travel distance using partial information is a challenging yet crucial 
 ingredient for the system's success. 

Approaches have been proposed to predict generic time-series
and many of them have capitalized on the phenomenon that for each individual there often exist 
re-occurring small fragments of time (which we call ``snippets'') in their histories. 
By detecting and reusing such snippets, we are able to
reconstruct a day with the elements from previous relevant days.
We show an example of snippet learning for daily traveling time prediction and the difficulties it faces by using a commuter's daily
routines. It is worth noting that throughout the entire paper, we assume that besides the time-series itself, no other support information such as locations are available to the prediction algorithm. For example, to predict a day's travel distance, the algorithm's only input is a partial time-series of the distances traveled in each interval. With a 30-minutes interval, the whole day will have 48 time-series entries, and we aim to use the first half of them to predict the accumulated travel distance for the whole day.

Imagine that a person in our example has
two usual routines: 1) on workdays the person goes to
work by a particular bus line that stops outside the apartment every
8 a.m., and arrives at the workplace around 9 a.m. The person gets
lunch around 12 p.m. at someplace near the workplace everyday, and
finishes work around 5 p.m., 2) on weekends the person prefers going to 
the beach in the morning and
coming home in the evening. In the ideal case, the person begins and 
finishes the same activity at the exact
 same time on every workday, and the resulting time-series 
 for travel distances would be identical across days. With snippets, a
time-series for a workday would then be transformed into a series of snippets like $<$$\Delta d_{walking ~to~ bus~ stop~ A}$,
$\Delta d_{10km~ ride ~on ~bus~ line~1}$, $\Delta d_{walking ~into ~office}$, $\Delta d_{working}$, ...$>$. 
 Now to predict how much 
 further the moving object will move for the
 remainder of the day at a certain time on the day (e.g. midday), 
we are left with a simple task. For every interval of the snippet sequence in the example,
if the current day shows an identical partial time-series for that interval, 
the person is likely to be working that day and is likely to yield the same total travel distance as any other workday. 
The same method works for the weekends too.

In reality, such patterns do repeat themselves, 
only not in such a perfectly aligned way but
 instead often on a shifted timeline and at a differing pace.
Instead of having high coherences at all times between two working
days of a person, in reality a day's time-series may often be partially similar to
and partially divergent from another day's, posing a serious challenge
for the aforementioned prediction method. There are
many possible causes which prevent a perfect resembler for a 
snippet sequence from happening. For example, the bus in the morning may
be 20 minutes late, or the person may wait for a coffee to miss
the bus he/she is supposed to take. Then, the person may have a later than usual lunch at work. Finally, the person on one day decides to 
do usual item A/B in the order of B/A. Coupled with the huge number of non-work-related locations 
a person could go to and the numerous possible sequences of visiting them, 
the resulting time-series could have a huge variety of distortions to the regular time-series. 
In such cases, how to effectively learn representative snippets and 
how to use them effectively remains a major challenge.

\eat{
To solve this complex problem, we take a step
forward and observe that it is
most beneficial if we put all the days' time-series together and discover the
significant structural elements (i.e. snippets) which could together roughly resemble most of the historical days. A
time-series would then be transformed into a series of snippets like $<$walking to bus stop A,
10km ride on bus line 1, walking into office, working at office,
walking to restaurant, having lunch, ...$>$. Ideally, with sufficient
historical data, an individual's daily travel patterns can be
reconstructed sparsely and accurately from the individualized
snippets. That is, most of the training time-series can be reconstructed
from only a small number of snippets with a small error. Naturally, 
the more snippets two days share, the more likely they
will have similar end-of-the-day travel distances. \eat{Moreover, instead
of using one layer of abstraction (finding the snippets) on the
time-series, we show that by repeating the snippet learning procedure,
we are able to find even more abstract snippets, e.g. the previous
snippets now becomes $<$commuting to work, working, lunch activity,
...$>$, further improving the abstraction of the reconstruction and hence
the prediction accuracy}.}

To solve this complex problem, we adopt the concept of snippets but take a step forward and propose a robust learning and time-series prediction model to systematically 
reduce the effect of such distortions. Specifically, we make the following contributions in this paper:
\begin{itemize}
\item We propose a novel regression model, which is based on convolutional neural networks, to solve the robust snippets learning and periodical time-series prediction problem. 
\item We propose a novel technique called temporal embedding to improve the classical convolutional neural networks' capability for learning robust snippets and for predicting accurately. We design a network layer based on this concept, devise a complete four layer network (TeNet) for regression, and solve the corresponding backpropagation problem. We also offer a detailed case study to illustrate the effect of temporal embedding.
\item We conduct extensive experiments on 15 individual datasets representing three data modalities and one synthetic dataset to evaluate the advantages and characteristics of the proposed model.
\end{itemize}

The rest of the paper is organized as follows. Next in Section \ref{sec:pf} we present the background and relevant literature of the problem studied. In Section \ref{sec:model} we give the intuition behind TeNet, describe in detail the technique of temporal embedding and other layers of TeNet, and offer solutions to the backpropagation of TeNet. We then enter Section \ref{sec:exp} and evaluate the proposed model. Finally we conclude our work in Section \ref{sec:conc}.

\section{Background and Related Work}
\label{sec:pf}
Learning abstract features (with neural networks in many cases) has been extensively
studied in recent years and has proved effective in many applications. For instance, numerous studies \cite{bengio2007greedy,DBLP:journals/ftml/Bengio09,DBLP:journals/jmlr/LarochelleBLL09,lee2009convolutional,DBLP:journals/pr/FischerI14} have shown that deep neural networks perform well for complex computer vision classification tasks, while many demonstrate that success can be achieved with deep learning architectures for audio classification tasks as well\cite{lee2009unsupervised,DBLP:conf/icml/NgiamKKNLN11}. These well-performing deep neural networks have a variety of core ideas, ranging from restricted boltzmann machines that utilize an energy model \cite{DBLP:series/lncs/Hinton12, DBLP:conf/nips/LeeEN07, DBLP:series/lncs/Hinton12}, to sparse autoencoders that introduce an unsupervised ``denoising'' mechanism to remove insignificant, noisy signals from data \cite{DBLP:conf/icml/VincentLBM08,bengio2007greedy,DBLP:journals/jmlr/VincentLLBM10}, to using convolution as an effective way to learn representative features robust to geometric locations of images \cite{lee2009convolutional, DBLP:conf/ijcai/CiresanMMGS11}.

The main advantage of such methods is that they have a strong capability of 
unravelling the hidden hierarchical structure of data to derive representative 
features. Moving from a shallower architecture to a deeper architecture, these models progressively
 detect essential components of the data from local parts like 
 strokes in human handwriting, to global compositions such as digits 
 or objects. Among the variations of neural networks, inspired by biological processes~\cite{DBLP:journals/nn/MatsuguMMK03}, convolutional networks in particular excel in finding such abstract features
 that are robust to geometric variations in images \cite{lee2009convolutional}. 
 Interestingly, such advantages of convolutional neural networks are present not only in vision tasks,
 but also in speech recognition \cite{DBLP:journals/taslp/Abdel-HamidMJDPY14,DBLP:conf/icassp/DengLHYYSSZHWGA13,DBLP:journals/corr/HannunCCCDEPSSCN14} and natural language processing \cite{collobert2008unified,DBLP:journals/jmlr/CollobertWBKKK11}. 

\eat{Often referred to is the abstract feature learning example in computer
vision. In this scenario, objects are often
characterized by their edges. Naturally, with the countless objects in
the real world, it is not surprising that there are some shapes of
edges are more common on a particular type of object, while some
shapes may very likely be present on another. These shapes of edges
may not be associated with particular meanings, however they do
capture the structure of the real world: a real-world object is often
composited from such distinctive elements.}

Now we consider the periodical time-series prediction problem for
data such as daily traveling distances or daily household power
consumptions. To tackle this problem,  conventionally statistical models such as autoregression and its variants are strongly favored. 
While in the past decade, realizing there is abstract and structural information 
beneath the raw numeric values in the time-series, researchers have 
experimented to discover such patterns by clustering or ``motif'' discovery 
\cite{DBLP:conf/kdd/RakthanmanonCMBWZZK12,schneider2013unravelling, schneider2013daily}. Though conceptually similar, these ``motifs'' usually are concrete subsequences that are restricted by specific mathematical definitions, 
which differentiate themselves from the concept of abstract, representative snippets in our paper. 
However, how to design a method that can find abstract patterns as well as predict future values, that meanwhile is
 robust to various temporal distortions and misalignment, is yet to be answered.  Inspired by the success of convolutional neural networks, 
 we investigate using convolution-based neural networks to address this problem.

\begin{figure*}[t]
\vspace{-1.3cm}
\centering
%\vspace{-0.1cm}
\includegraphics[scale = 0.55]{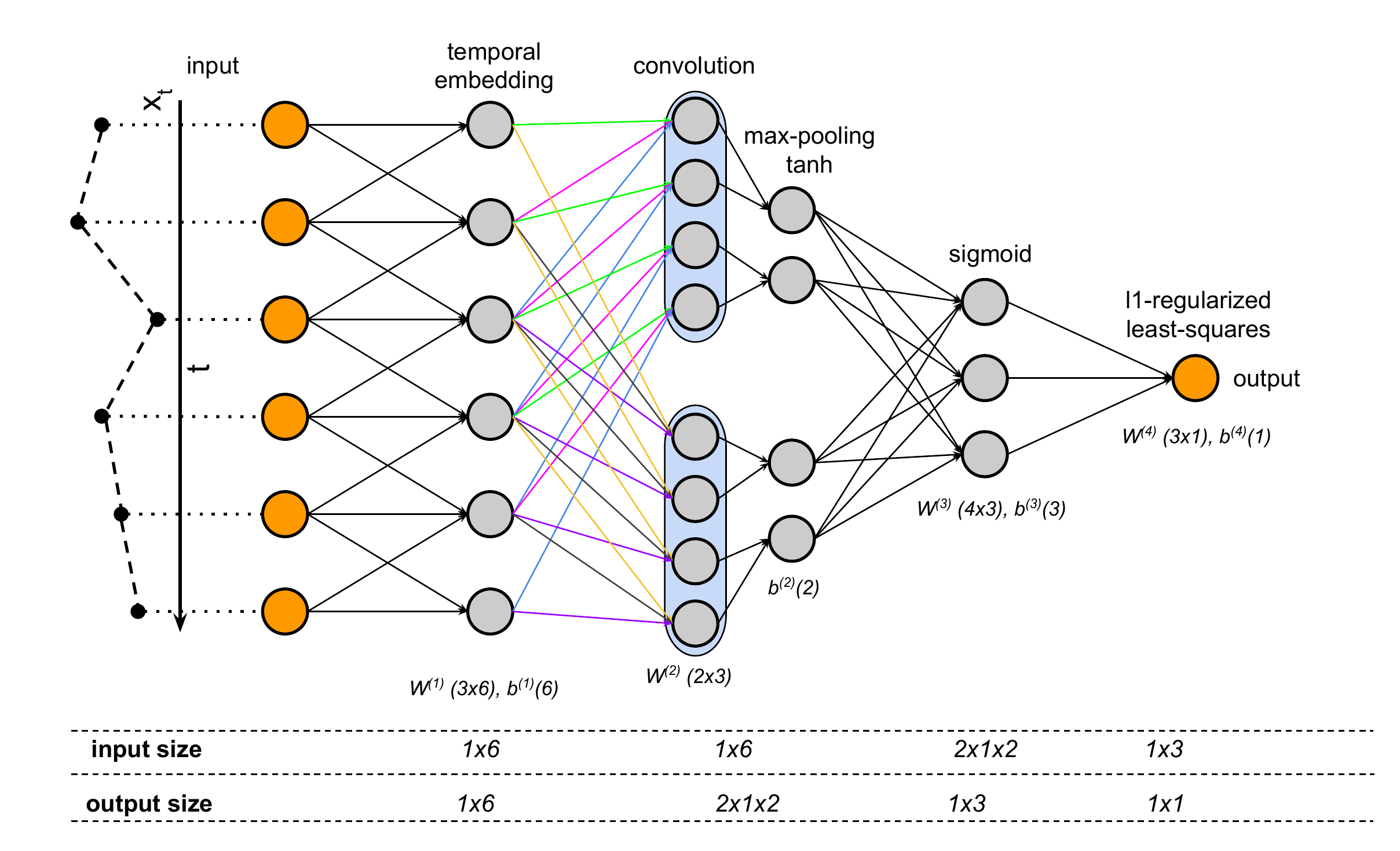}
\vspace{-0.6cm}
\caption{\scriptsize{An instance of TeNet for 6-d input. It is composed of a sparsely connected temporal embedding layer, a convolution/pooling layer with two filters of size $1\times3$ and pool size $1\times2$ (following the conventions in constructing convolutional neural networks, the convolution layer and max-pooling layer are illustrated as a single layer), a fully-connected sigmoid layer that transforms the feature map from size $2\times1\times2$ into $1\times3$, and finally an l1-regularized least-squares regression layer that yields the predicted value. $W^{(l)}, b^{(l)}$ are the weights and bias of the connections between layers $l$ and $l+1$. Connections with the same colors in the convolution layer indicate that those connections share the same weights, and the two shaded areas represent the two feature maps from the filters. The dimensionalities of the weights, the input and the output for each layer are provided at the bottom. Biases are not illustrated in this figure.}}
\label{fig:cnn}
\end{figure*}

\section{The Model}
\label{sec:model}

\subsection{Intuition}
The two main challenges for the periodical time-series prediction are: 1) how to find representative snippets for the prediction of future changes; and 2) how to minimize the effect of distortions in the temporal domain and get accurate regression results. Here we examine the two challenges separately and propose solutions to them from a neural networks perspective.

%\vspace{-0.4cm}
The first challenge, i.e. snippet learning, involves finding abstract sequences in the training time-series. Naturally there is an assumption that the snippets should only be of moderate length. For example, if we were to predict daily human mobility, a time window of from one half-hour to  a few hours would be a reasonable setting, as intuitively such a period of time should be enough to cover most of the common trips in daily life. Hence in the prediction model, we examine such periods of time using a convolutional approach. We create randomly initialized filters that have a given, moderate length as the length of the target snippets. In 2D image classification tasks, filters in convolutional neural networks are often used as edge detectors, while in ours, the filters will act as ``snippet detectors''. In the training phase, the weights for the filters will be adjusted during the backpropagation so that they respond maximally to the reoccurring and significant components in the training data.

We then solve the second challenge by adding a ``temporal embedding'' operation in the neural network. The temporal embedding process provides a supervised way of denoising subspace learning. When dealing with time-series, a na\"{i}ve technique is to ``shift'' the training data forward and backward along the timeline. For example, a shifting routine with windows size 1 would transform a training sample $x=<x_1,x_2,...,x_d> \rightarrow y$ into three training samples $x=<0, x_1,x_2,...,x_{d-1}> \rightarrow y, x=<x_2, x_3,...x_{d}, 0> \rightarrow y, x=<x_1,x_2,...,x_d> \rightarrow y$. Though useful sometimes, this na\"{\i}ve approach introduces heavy noise by including artificial training samples that may never actually happen in the real world. Also it is unable to benefit case where the order of the subsequence is changed. We argue that the na\"{\i}ve technique can evolve to a much more effective approach called temporal embedding that integrates into the learning process mechanisms for removing distortions. With temporal embedding,  two temporally-shifted copies are created for each sample during the learning process, and then the original sample and the two shifted copies are encoded into a single sample so that the processed sample will not only carry its own information, but also bear a piece of information for each of its shifted neighbors. Again, the weights for the encoding are learned in a supervised way during backpropagation.

Next we present an overview of the TeNet model.

\subsection{Model Overview}
We propose a convolutional neural network to learn the snippets from the periodic time-series as illustrated in Figure \ref{fig:cnn}. 
The model has three invisible layers, namely the temporal embedding layer, the convolution/max-pooling layer, and the sigmoid layer. The output layer is an l1-regularized least squares regression layer. The illustrated model is an example instantiation of the proposed model, with the input size, embedding window size, number of snippets, snippet size, max-pooling and sigmoid layer sizes to be 6, 1, 2, (1,3) and (1,2) and 3 respectively. The model implements the following work flow:

\begin{enumerate}
\item It takes an input sample, and applies the temporal embedding. This layer transforms the sample into a denser representation with not only the sample itself but also information of its potential temporal neighbors. The weights of the transformation are iteratively updated during the training process.
\item The embedded input is sent into a convolution layer where a set of filters, or snippet detectors, scan through the sample using the convolution operator. Each snippet will be convolved against the sample, resulting in a feature map considered as the snippet's response to that sample.
\item The snippets' responses to the sample, being supposedly sparse and representative, are input into a sigmoid layer to combine some of the responses into higher-level and more abstract representations in lower dimensions. This transformation also involves a set of weights that is learned over the training process.
\item Finally the abstract representation of the sample is used to perform an l1-regularized least-squares regression to obtain the predicted value. The intuition behind the l1 regularization is that if we consider the previous layer's output, ie. the high-level neuron's responses to the sample, as high-level pattern recognizers responses to the signal,  a sparse solution will utilize the most significant responses and hence will be less sensitive to noise~\cite{ng2004feature,schmidt2005least}. 
\end{enumerate}

In the following subsections we discuss the layers separately in detail. In the rest of the paper, the technical details of the neural network will be described mostly in vector forms, and we will use the assumptions and notations listed in Table \ref{tbl:not}.
\begin{table}[htp]
\centering
\vspace{-0.4cm}
\caption{Table of Notations}
\label{tbl:not}
\small{
\begin{tabular}{ | c |l| }
\hline  \textbf{Notation} & \textbf{Description} \\
\hline $x\in \mathbb{R}^d$ & the input time-series of length $d$ \\
\hline $l$ & the layer number\\
\hline $W^{(l)}$ & the weights for the $l^{th}$ layer\\
\hline $b^{(l)}$ & the bias for the $l^{th}$ layer\\
\hline $a^{(l)}$ & the input of neurons in the $l^{th}$ layer\\
\hline $z^{(l)}$ & the intermediate values for the $l^{th}$ layer\\
\hline $g^{(l)}()$ & the activation function for the $l^{th}$ layer\\
\hline $\delta^{l}$ & the intermediate error (cost) of the $l^{th}$ layer\\
\hline $J(W,b;x,y)$ & the network's cost given $W,b;x,y$\\
\hline $^T$ & the transpose operator\\
\hline $\cdot$ & the dot product operator\\
\hline $\odot$ & the element-wise product operator\\
\hline $*$ & the convolution operator\\
\hline $g^{\prime(l)}$ & the derivative of function $g^{(l)}$\\ 
\hline
\end{tabular}
}
\end{table}

\subsection{Temporal Embedding}
\label{ssec:te}
The temporal embedding layer aims to align less dominant samples to the dominant patterns by reducing the temporal distortions and misalignment (e.g. shifting or skewed sequence of events), corresponding to two cases in our previous example: 1) the commuter starts the day 30 minutes earlier than usual, so every event in the morning rush hour is shifted ahead equally by 30 minutes , 2) for some reason the commuter does not take the usual bus line which directly stops at his workplace, instead he/she takes a train and walks 1km to work from the station. In the resulting time-series we will see two distinct effects as a result of 1) and 2). For example, assume that on normal day the travel distance time-series segment in the morning will be $v=<0,1,2,4,1,0>$, then for case 1 we will have $u=<1,2,4,1, 0,0>$, and in case 2 it will be $u=<0,1,4,2,1,0>$. Now we assume both cases happen on the same day, giving us $u=<1,4,2,1,0>$, which is heavily distorted from $v$. It is a significant challenge for a prediction algorithm to realize that for the two days the travel distances should be very similar despite the sequences and the values of their time-series are so different. 

Temporal embedding addresses this issue, by optimally embedding a value's temporal neighbors into itself, so that for the whole dataset the dominant pattern remains unchanged but the distorted patterns are realigned. The layer is configured by one hyperparameter $d^{te}$ that controls how many neighbors of an element in each direction should be embedded to the element itself (the embedding size). This layer has $2\times d^{te}+1$ sets of parameters, represented by matrices $W^{(1)}_{l_j}, W^{(1)}_{m}$ and $W^{(1)}_{r_j} \in \mathbb{R}^{d^{(1)}\times d^{(1)}}$ , and the same number of constant sparse matrices $\widetilde{W}_{l_j},\widetilde{W}_{m}$ and $\widetilde{W}_{r_j} \in \mathbb{R}^{d^{(1)}\times d^{(1)}}$. The subscriptions $l$ and $r$ represent the direction of the neighbors on the timeline, and $j$ here means the weights for the $j^{th}$ neighbor in the final embedding. In the case of $d^{te}=1$, there are three $W$ matrices and three $\widetilde{W}$ matrices in this layer. The six matrices together implement the embedding operators. Here we use the input dimensions in Figure \ref{fig:cnn} (where $d^{(1)}=6$) as an example for how this layer works. 

The constant matrices, are defined as:
\begin{eqnarray}
\widetilde{W}^{(1)}_{l_1} = \left[ \begin{array}{cccccc}
0 & 0 & 0 & 0 & 0 & 0 \\
1 & 0 & 0 & 0 & 0 & 0 \\
0 & 1 & 0 & 0 & 0 & 0 \\
0 & 0 & 1 & 0 & 0 & 0 \\
0 & 0 & 0 & 1 & 0 & 0 \\
0 & 0 & 0 & 0 & 1 & 0 
\end{array} \right] \\
\widetilde{W}^{(1)}_{r_1} = \widetilde{W}^{(1)T}_{l_1}, \widetilde{W}^{(1)}_{m}=I=eye(d^{(1)})
\label{eqn:cons}
\end{eqnarray}

Weights in $W^{(1)}_{l_1}$,$W^{(1)}_{m}$  $W^{(1)}_{r_1}$ that correspond to the $1$s in $\widetilde{W}_{l_1}$, $\widetilde{W}_{m}$ and $\widetilde{W}_{r_1}$ represent the weights for the embedding of the sample's left neighbor (forward), the sample itself and its right neighbor (backward) respectively, and they are initialized with corresponding constant matrices respectively. The layer's output is subsequently defined as follows:
%\begin{flalign}
\begin{gather}
%\nonumber g^{(1)}(z^{(1)}) = b^{(1)} + z^{(1)} \cdot \frac {\left[ (W^{(1)}_{l1} + W^{(1)}_{r:1})\odot \widetilde{W}^{(1)} + I \right]}{2\times d^{(1)} + 1}
z^{(1)} = a^{(1)} \cdot(  W^{(1)}_{l_1} \odot \widetilde{W}_{l_1} + W^{(1)}_{r_1} \odot \widetilde{W}_{r_1} + W^{(1)}_{m} \odot \widetilde{W}_{m}) +b^{(l)}\\
g^{(1)}(z^{(1)}) = z^{(1)}
\end{gather}
%\end{flalign}
$\widetilde{W}^{(1)}$ enforces a constraint that the connections between this layer and its input are restricted, and only the weights at the desired neighboring positions for each element are used in the final embedding for that element. The layer yields the temporal embedded output $g^{(1)}(z^{(1)})\in \mathbb{R}^{d^(1)}$, or $6$ in this example, as the output of the layer. One can also use the sigmoid function as the activation function in the temporal embedding layer, though our experiments show that the difference it makes on the prediction accuracy is insignificant (most of the times adding the sigmoid activation will slightly decrease the prediction accuracy).

The layer's output is a vector of the same size as the input, however the embedded sample is now significantly more robust to temporal distortions. With temporal embedding, the model detects dominant patterns in the training time-series, and tries to correct the systematical distortions within the specified time window. Using the commuter example, the model will find that the person's regular time for the bus to work, and will try to realign the systematical misalignment on those unusual days. Some readers may argue that a simple moving average algorithm might be able to solve the distortion problem; however temporal embedding is far more effective, as the concrete example below shows.

\paragraph{Discussion and Case Study} Recall our example with $v$ and $u$, where $v$ represents the dominant pattern in the dataset, while $u$ represents a day that in fact will yield a similar end-of-day result but shows very distorted patterns in its time-series. Now given the parameter matrices $W_l, W_m, W_r$ and the constant matrices $\widetilde{W}_l,\widetilde{W}_m,\widetilde{W}_r$ initialized as in Equation \ref{eqn:cons}, our objective is to realign $u$ with $v$ by eliminating the distortion, and meanwhile keeping $v$ as unchanged as possible, which is effectively equivalent to solving the following minimization problem in Equation \ref{eqn:eq}:
\begin{eqnarray}
&&v^t = v \cdot(  W_{l} \odot \widetilde{W}_{l} + W_{m} \odot \widetilde{W}_{m} + W_{r} \odot \widetilde{W}_{r} ) +b \label{eqn:at}\\
&&u^t = u \cdot(  W_{l} \odot \widetilde{W}_{l} + W_{m} \odot \widetilde{W}_{m} + W_{r} \odot \widetilde{W}_{r}) +b \label{eqn:bt}\\
&&\arg\min_{W_l,W_r}{ ||v^t-v||^2}+ || u^t -v||^2 \label{eqn:eq}
\end{eqnarray}
where $v^t$ and $u^t$ are the embedded new time-series. By solving the optimization, the non-zero weights in $W_l\odot \widetilde{W}_l$,  $W_m \odot \widetilde{W}_m$ and $W_r \odot \widetilde{W}_r$ are determined as $<0, 0.61 ,$\\$0.24,0.44,1>$, $<0,-0.22,0.4,-0.15>$ and $<0.66,0.24,2.1,$\\$1>$ respectively. Now $v^t$ and $u^t$ can be calculated according to Equations \ref{eqn:at} and \ref{eqn:bt}, and we subsequently investigate how temporal embedding performs in terms of preserving $v$ and realigning $u$ to $v$, compared with the moving average approach, with $v^s$ and $u^s$ being the output of $v$ and $u$ of a moving average of window size 3 ($v_i = \overline{<v_{i-1},v_i,v_{i+1}>}$). 

\begin{table}[htp]
%\centering
\vspace{-0.5cm}
\caption{Temporal Embedding vs. Moving Average}
\hspace{-0.5cm}
\begin{tabular}{ | c |l|c|l| }
\hline $v$ & $<0,1,2,4,1,0>$ & $u$ & $<1,4,2,1,0,0>$\\
\hline  $v^t$ &  $<0,1,2,4,1,0>$& $u^t$ & $<0,1,2,4,1,0>$ \\
\hline $v^s$ & $<0.5,1,2.3,2.3,1.7,0.5>$ & $u^s$ &  $<2.5,2.3,2.3,1,0.3,0>$\\
\hline
\end{tabular}
\begin{tabular}{ |c|c|c|c| }
\hline & \textbf{Squared Error} & \textbf{Intersection} & \textbf{Pearson's}\\
\hline $v, u$ &  4.5&4&0.11\\
\hline $v, v^t$ &  0&8&1\\
\hline $v^t, u^t$ & 0 &8&1\\
\hline $v, v^s$ & 2 & 6.3&0.87\\
\hline $v^s,u^s$ & 3.1 & 5.2&0.02\\
\hline
\end{tabular}
\label{tbl:comp}
\end{table}

Table \ref{tbl:comp} measures the relations between the vectors before and after the transformations with three metrics, namely squared error, intersection and Pearson's correlation. First we note that $u$ is so distorted that the correlation between $v$ and $u$ is merely $0.11$, which can be considered ``uncorrelated''. Now we examine the differences between the effects of temporal embedding and moving average.

Ideally, the transformation should show the following properties: 1) since $v$ represents the reoccurring pattern in the training set, we want $v^t$ to be as unchanged as possible after the transformation 2) after the transformation, $u^t$ should be as similar to $v^t$ as possible, indicating that the misalignments in $u$ has been minimized and $u$ is realigned to the representative sample $v$. We verify the two aspects by examining the relations between $v$ and $v^t$, and that between $u^t$ and $v^t$, and observe that temporal embedding has achieved both goals.

First we observe that $v^t$ is identical to $v$ (with $0$ squared error), while $u^t$ has been transformed to a form that is perfectly identical to $v$ and $v^t$ now, with the dominant values at the second and third positions swapped and realigned to the third and forth position to be more inline with $v$. However, we can see moving average resulted in a squared error of $2$ between $v$ and $v^s$, showing that $v$ has not been preserved successfully in the transformation. Second, though moving average does strengthen the relation between $v$ and $u$ by reducing the squared error ($4.5 \rightarrow 3.1$) and by increasing the similarity by intersection, it has even resulted in a drop in the correlation ($0.11\rightarrow 0.02$ compared with the original $v$ and $u$). We conclude its result is clearly less successful compared to temporal embedding ($4.5 \rightarrow 0$ in squared error, $4\rightarrow 8$ in intersection, and  $0.11\rightarrow 1$ ). 

It is worth noting that although the temporal embedding layer in the proposed neural network is not exactly the same as in Equation \ref{eqn:eq} as it does not have knowledge initially about which samples hold the representative patterns, as  the training proceeds, the weights will progressively favor the reoccurring patterns, and eventually approach the solution of Equation \ref{eqn:eq}. Next we describe the convolution, the max-pooling and the sigmoid layers.

\subsection{Convolution, Max-pooling and Sigmoid}
The convolution/pooling layer performs a series of discrete 1-d convolutions $W^{(2)}_i*a^{(2)}$ with a specified number of filters $n^f$ of a specified length $d^f$. Each of the filters ``sweeps'' through the entire input signal and  takes the input signal segment at the corresponding position as input. With a filter kernel $W^{(2)}_i=<W^{(2)}_{id^f},W^{(2)}_{id^f-1},...,W^{(2)}_{i1}>$ (taking the convention of reversely-ordered weights for convolution kernels and outputs), the $i^{th}$ filter's output has the $k^{th}$ element:
\begin{eqnarray}
(W^{(2)}_i*a^{(2)})[k] = \sum^{d^f}_{p=1} W^{(2)}_{ip}  a^{(2)}_{k+p-1}
\label{eqn:filter}
\end{eqnarray}

In the example in Figure \ref{fig:cnn} we have set two filters with size 1x3, hence in the convolution layer, each neuron will only be connected to three neurons from the temporal embedding layer. Such sparse connectivity between the filters to their inputs enforces that the convolution layer will be focusing on finding the local snippets with moderate lengths.

Though the convolution traverses the entire time-series in a sliding-window style and seemingly has a positive effect in reducing the temporal distortions, it is very different from temporal embedding. The main factor differentiating them is in the weight-sharing scheme (see Figure \ref{fig:cnn}). A filter in the convolution layer has its weights shared among all its output neurons (meaning a filter is sliding through the data, trying to match the same particular pattern), while in temporal embedding each neuron has individualized weights to enable optimal local embedding for each position. Such flexibility enables it to identify and realign much more complex distortions and misalignments. For example, given $v=< 0,1,2,4,1,0 >$, convolution will not be able to recognize the close relation between $u =< 1,4,2,1,0,0 >$ and $v$ because of the heavy distortions in both the positions and the sequences. In the experiments we will also show that without the temporal embedding layer, convolutional neural network does not work well on such time-series.

The output of the convolution will be of the size $n^f \times d^{(-d^f+1)}$. In Figure \ref{fig:cnn}'s example where $n^f=1,d=6,d^f=3$, we have the 8 neurons in the convolution layer. The output is then received by the max-pooling layer, where only the maximal value is kept from any pool of $1 \times 2$. The filter's output will hence be  down sampled and transformed by an element-wise hyperbolic tangent function, reducing the output to 4-dimensional. Then as the last hidden layer, the sigmoid layer will perform a projection from the convolution/pooling's output to a further reduced dimension as a means of both learning non-linear features and dimension reduction. Finally, the input is transformed into a dense, robust and representative feature representation of $1\times 3$. Intuitively we can consider the sigmoid layer as a higher-level feature learner, after the convolution layer has discovered those relatively more ``local'' snippets.

\subsection{$l1$-regularized Least-squares}
The output layer of the proposed model is a l1-regularized least-squares regression layer, defined as:
\begin{eqnarray}
g^{(4)}(a^{(4)}) = a^{(4)} \cdot W^{(4)T} + b^{(4)}
\end{eqnarray}
with the cost function in the from of:
\begin{eqnarray}
J(W,b;x,y) = \frac{1}{2}||a^{(4)} \cdot W^{(4)T} + b^{(4)} -y ||^2 + \lambda||W^{(4)}||_1
\end{eqnarray}
where $\lambda$ is a hyperparameter for the weight of the regularization term. 

The advantage of using the $l1$ regularizer over $l2$ is that the $l1$ regularizer forces the optimization to find a sparse solution that only uses the most distinctive high-level features to conjure the final prediction \cite{ng2004feature,schmidt2005least}. With the $l2$ regularizer the weights tend to have smaller variance, often making the model spread the energy thinly across all  features, hence making the model less distinctive and less accurate.

\subsection{Backpropagation}
The parameters in the network are updated by stochastic gradient descent. In particular, $W^{(4)}$ can be learned by:
\begin{eqnarray}
\nonumber \hspace{-0.4cm} \frac{\partial J(W,b;x,y)}{\partial W^{(4)}} = a^{(4)T}(a^{(4)} \cdot W^{(4)T} + b^{(4)} -y) + \lambda~sign(W^{(4)T})
\end{eqnarray}
Where $sign()$ is the sign of a vector. One can speed up this optimization process using the methods proposed in \cite{DBLP:conf/acl/TsuruokaTA09}.

To update the parameters in the temporal embedding layer, taking $W^{(1)}_{l}$ as an example, we apply the chain rule and arrive at:
\begin{gather}
\nonumber \frac{ \partial J(W,b;x,y) }{\partial W^{(1)}_{l_1}} = \frac{ \partial J(W,b;x,y) }{\partial g^{(1)} } \odot\frac{ \partial g^{(1)} }{\partial z^{(1)} }  \odot\frac{ \partial z^{(1)} }{\partial W^{(1)}_{l_1}}\\
=\frac{ \partial J(W,b;x,y) }{\partial z^{(1)} } \odot\frac{ \partial z^{(1)} }{\partial W^{(1)}_{l}} = \delta^{(1)}_l \odot \frac{ \partial z^{(1)} }{\partial W^{(1)}_{l}}
%\delta^{(1)} &=& flip(W^{(2)})*\delta^{(2)}\\
%\frac{ \partial J(W,b;x,y) }{\partial W^{(1)}_{l}} &=& z^{(1)} \cdot  (flip(W^{(2)})*\delta^{(2)}) \label{eqn:grad}
\end{gather}
Since the element-wise product has the property:
\begin{eqnarray}
a^{(1)}\cdot(W^{(1)}_{l} \odot \widetilde{W}^{(1)}_{l}) = (a^{(1)} \odot \widetilde{W}^{(1)}_{l}) \cdot W^{(1)}_{l} 
\end{eqnarray}
we have the partial derivative of $z^{(l)}$ w.r.t. $W^{(1)}_{l}$ as:
\begin{eqnarray}
\frac{ \partial z^{(1)} }{\partial W^{(1)}_{l}} =  a^{(1)}\odot\widetilde{W}^{(1)}_{l} 
\end{eqnarray}
We calculate the error propagates from layer 2 to layer 1 as:
\begin{gather}
\nonumber \delta^{(1)} = \frac{\partial J(W,b;x,y)}{\partial z^{(2)}} \odot \frac{\partial z^{(2)}}{\partial g^{(1)}} \odot \frac{\partial g^{(1)}}{\partial z^{(1)}}  \\
= \sum_{i=1}^{d^f}\delta_i^{(2)} \odot \frac{ \partial W^{(2)}_i * a^{(2)}} {\partial a^{(2)}}
= \sum_{i=1}^{d^f}\delta^{(2)}_i *flip(W^{(2)}_i)
\label{eqn:d1d2}
\end{gather}
where $flip()$ returns the input vector in reversed order. With the convolution layer's back propagated error being $\delta^{(2)}$ (which can be calculated by the method described in \cite{726791}),$\frac{ \partial J(W,b;x,y) }{\partial W^{(1)}}$ can therefore be updated with the gradient:
\begin{gather}
\hspace{-0.5cm}
\frac{ \partial J(W,b;x,y) }{\partial W^{(1)}_{l}} = \sum_{i=1}^{d^f}  \left[  \delta^{(2)}_i*flip(W^{(2)}_i)\right] \odot ( a^{(1)}\odot\widetilde{W}^{(1)}_{l} )
\label{eqn:upd}
\end{gather}
$W^{(1)}_{l}$ and $W^{(1)}_{m}$ can be updated using similar procedures. Meanwhile, $b^{(1)}$ is updated with the gradient:
\begin{eqnarray}
 \frac{ \partial J(W,b;x,y) }{\partial b^{(1)}} =   \sum_{i=1}^{d^f} \delta^{(2)}_i
\end{eqnarray}

Next we present the experimental results and offer in-depth analysis and discussion.

\begin{table*}[htp]
\vspace{-0.8cm}
\scriptsize
      \caption{Performance comparison }
      \label{tbl:comp}
  \centering
    \begin{tabular}{c|c|c|c|ccccc|ccccc}
    \toprule
        \multirow{2}{*}{ID}    & \multirow{2}{*}{n} & \multirow{2}{*}{[\texttildelow]} & \multirow{2}{*}{$\sigma$}   &  \multicolumn{5}{c}{HitRate (\%, @20\%|@30\%)}  & \multicolumn{5}{c}{Error (MRE/MSE)} \\
	& & & & TeNet & SVLN & SVSIG & SVPOLY & MKR        & TeNet & SVLN & SVSIG & SVPOLY & MKR \\
    \midrule
\midrule
    \multicolumn{14}{c}{Household Power Consumption - Australia (HPC-AU)} \\
\midrule
8&874&3.1-36.4&5.6&\textbf{\textcolor{magenta}{71}}|\textbf{\textcolor{blue}{89}}&55|74&59|83&60|82&48|75&
\textbf{\textcolor{red}{0.16}}|\textbf{\textcolor{black}{7.7}}&0.24|14.4&0.23|10.5&0.23|9.5&0.29|17.9\\

15&870&1.5-30.9&5&\textbf{\textcolor{magenta}{65}}|\textbf{\textcolor{blue}{84}}&58|78&56|79&59|80&53|75&
\textbf{\textcolor{red}{0.2}}|\textbf{\textcolor{black}{11.2}}&0.2514.2&0.26|15&0.24|13.6&0.26|18.5\\

14&670&1.7-59.2&7.2&\textbf{\textcolor{magenta}{65}}|\textbf{\textcolor{blue}{79}}&42|64&40|60&50|72&63|77&
\textbf{\textcolor{red}{0.21}}|\textbf{\textcolor{black}{10}}&0.26|13.7&0.31|15.3&0.24|11.1&0.31|\textbf{\textcolor{black}{10}}\\

7&665&8.4-38.4&5.0&\textbf{\textcolor{magenta}{75}}|\textbf{\textcolor{blue}{92}}&67|88&73|90&\textbf{\textcolor{magenta}{75}}|91&72|89&
\textbf{\textcolor{red}{0.14}}|15.9&0.16|17.1&0.15|15.8&\textbf{\textcolor{red}{0.14}}|14.4&0.16|\textbf{\textcolor{black}{15.7}}\\

5&661&0.2-8.0&1.6&\textbf{\textcolor{magenta}{71}}|\textbf{\textcolor{blue}{82}}&58|76&57|76&58|76&70|\textbf{\textcolor{blue}{82}}&
0.84|1&1.17|1.82&1.1|1.7&1.1|1.7& \textbf{\textcolor{red}{0.39}}|\textbf{\textcolor{black}{0.9}}\\

%\hline
12&243&7.6-27.2&2.8&\textbf{\textcolor{magenta}{90}}|\textbf{\textcolor{blue}{97}}&88|96&\textbf{\textcolor{magenta}{90}}|96&85|92&78|91&
\textbf{\textcolor{red}{0.09}}|\textbf{\textcolor{black}{4.6}}&0.1|\textbf{\textcolor{black}{4.6}}&\textbf{\textcolor{red}{0.09}}|4.8&0.12|8.2&0.17|8\\

10&242&4.3-42.6&7.7&\textbf{\textcolor{magenta}{84}}|\textbf{\textcolor{blue}{96}}&74|92&78|94&70|90&71|83&
\textbf{\textcolor{red}{0.12}}|8&0.17|7.5&0.13|9.2&\textbf{\textcolor{red}{0.12}}|10.6&0.16|\textbf{\textcolor{black}{7.15}}\\

1&241&8.9-37.9&4.6&\textbf{\textcolor{magenta}{84}}|\textbf{\textcolor{blue}{96}}&80|95&75|92&76|91&76|93&
\textbf{\textcolor{red}{0.12}}|12&\textbf{\textcolor{red}{0.12}}|11.2&0.14|14&0.15|22& 0.13|\textbf{\textcolor{black}{9.23}}\\

13&241&4.4-46.7&6.3&\textbf{\textcolor{magenta}{65}}|\textbf{\textcolor{blue}{82}}&62|78&59|79&63|80&59|80&
\textbf{\textcolor{red}{0.19}}|22&0.2|22.3&0.2|26&0.21|23&0.24|\textbf{\textcolor{black}{18.7}}\\

29&233&17.3-73.5&11.3&\textbf{\textcolor{magenta}{77}}|\textbf{\textcolor{blue}{93}}&75|92&59|80&\textbf{\textcolor{magenta}{77}}|\textbf{\textcolor{blue}{93}}&76|92&
0.14|43&\textbf{\textcolor{red}{0.13}}|31&0.2|85.0&\textbf{\textcolor{red}{0.13}}|34&0.15|\textbf{\textcolor{black}{37.6}}\\
    \midrule
    \multicolumn{14}{c}{Household Power Consumption - France (HPC-FR)} \\
\midrule
    1     &   161 & 10-79.5    &  10.3      &    \textbf{\textcolor{magenta}{64}}|\textbf{\textcolor{blue}{83}}   &   56|75    &   53|73    &   60|75        &   63|81   &
      \textbf{\textcolor{red}{0.18}}|74.6     &  0.23|111     &  0.26| 110    &   0.22|100    &     0.2|\textbf{\textcolor{black}{67}}  \\
    \midrule
\multicolumn{14}{c}{Human Mobility - Traveling Distance (HMD) } \\
    \midrule
8&206&8-99&15.2&46|\textbf{\textcolor{blue}{63}}&\textbf{\textcolor{magenta}{48}}|62&35|50&\textbf{\textcolor{magenta}{48}}|\textbf{\textcolor{blue}{63}}& 33|45&
\textbf{\textcolor{red}{0.28}}|\textbf{\textcolor{black}{169}}&0.29|170&0.35|198&0.31|300& 0.4|323\\
12&156&9.5-60&11&\textbf{\textcolor{magenta}{59}}|\textbf{\textcolor{blue}{83}}&56|77&43|65&54|70&45|67&
\textbf{\textcolor{red}{0.20}}|\textbf{\textcolor{black}{74.5}}&0.23|101&0.28|96&0.27|285& 0.27|103\\
\midrule
\multicolumn{14}{c}{Human Mobility - Traveling Time (HMT)} \\
    \midrule
8&193&55-345&47&\textbf{\textcolor{magenta}{51}}|\textbf{\textcolor{blue}{70}}&47|64&42|58&40|57&45|66&
\textbf{\textcolor{red}{0.23}}|\textbf{\textcolor{black}{32.1}}&0.24|36.6&0.28|59&0.35|179& 0.25|43\\
12&243&37-280&32.4&\textbf{\textcolor{magenta}{61}}|\textbf{\textcolor{blue}{74}}&58|70&48|67&49|68&57|\textbf{\textcolor{blue}{74}}&
\textbf{\textcolor{red}{0.21}}|\textbf{\textcolor{black}{24.0}}&0.23|29.4&0.25|29&0.3|60& 0.22|25\\
\midrule
    \bottomrule
    \end{tabular}%
  \label{tab:res}% 
\vspace{-0.5cm}
\end{table*}

\vspace{-0.2cm}
\section{Experiments}
\label{sec:exp}
In the experiments, we conduct extensive tests on the proposed model, with 15 individual datasets and 4 competitive methods. The goals of the experimental studies are fourfold: 1) to evaluate the prediction performance of the proposed model, in terms of prediction accuracy, and compare it with the competitive models; 2) to evaluate the model's behavior and sensitivity to features of diverse datasets; 3) to investigate the isolated effects of temporal embedding; and 4) to visualize the snippets and show how they work with intermediate values from the learning process.

\subsection{Datasets}
To support the comprehensive evaluation, we use a variety of univariate, periodical time-series datasets that represent three modalities, ranging from human mobility patterns to household power consumption. The reason we choose these modalities is that the behaviors they represent are expected to exhibit complex periodical patterns in daily cycles, 
 which is an ideal testbed for the proposed model to demonstrate its capability of discovering and capturing such abstract features and to test its robustness to various factors.

The first modality is  Human Mobility - daily traveling Distance (HMD) in kilometers, and the second is Human Mobility - daily traveling Time (HMT) in minutes. Both modalities are extracted from the LifeMap \cite{yonsei-lifemap-2012-01-03}) that contains human mobility traces collected from eight individuals, spanning from a few months to around two years. In total there are  52,819 position fixations, most of which are from regular sampling every two to five minutes. HMD is the total displacement for an individual in a day, and HMT is accumulated from short-term movements calculated as follows: for each five minute interval, if the individual's displacement is higher than 500 meters~\footnote{median errors of localization with assisted GPS, WiFi positioning and cellular network positioning are reported to be 8, 74 and 600 m \cite{zandbergen2009accuracy}}, then the five-minute period is counted as ``traveling'' and is accumulated to the daily total traveling time. 

The third modality is daily Household Power Consumption (HPC). Two datasets are used for this modality, i.e. household power consumption datasets from France\footnote{\url{https://archive.ics.uci.edu/ml/datasets/Individual+household+electric+power+consumption}} and Australia\footnote{\url{http://data.gov.au/dataset/sample-household-electricity-time-of-use-data}} (HPC-FR, HPC-AU). HPC-FR consists of \\2,075,259 active power consumption in watt sampled every minute for 48 months from a single household. HPC-AU consists of $618,189$ household power meter readings in kwatt hour sampled every 30 minutes from $31$ households for up to 29 months.

To prepare the data, we developed a program to extract only the samples that have complete (or nearly complete) day cycles, meaning that every data sample used must have regular readings in each period of time in a complete day. To obtain meaningful results, only individuals  with more than $150$ days of records are used in the experiments. 

For the human mobility datasets, we use the two individuals' datasets with the highest quality of data in terms of timespan (>150 days) and sampling frequency.  We extract the traveling distances and traveling times for each interval (e.g. a 30 minutes interval creates 48-d time-series for a day), and use the resulting time-series for the experiments. Similar preprocessing is applied on the power consumption datasets. After preprocessing, each time-series sample has $d$ elements as $x=<x_1,...,x_d>$, each $x_i$ is the occurred value in the corresponding time interval (non-cumulative). 

For each individual dataset, we randomly divide the samples equally into three folds: the training set, the validation set and the test set. The model is trained using the training set, and is then tested on the validation set. Such cross-validation is performed on the same individual dataset for five times with random splits, and the reported performance is the averaged value cross the five iterations. The settings of hyperparameters with the best validation performance are kept as the hyperparamters of the model. Finally we test the model on the test set and report the performance.

%\textbf{\textcolor{red}{
%\textbf{\textcolor{black}{
%\textbf{\textcolor{blue}{
%\textbf{\textcolor{magenta}{

\subsection{Evaluation Settings}
\label{ssec:es}
For evaluation we consider the periodical accumulation prediction problem, where each input $x'\in \mathbb{R}^{d'} (d'< d)$ is a head segment of a complete $x$ and corresponds to a target value $y = \sum_{i=0}^dx_i$ representing the periodical accumulation. Clearly the model can be used to perform other types of prediction such as time-series forecast or $k$-ahead prediction. Due to space limit here we use periodical accumulation prediction as a showcase for TeNet's performance advantages. 

TeNet is implemented using Python with the Theano framework\footnote{\url{http://deeplearning.net/software/theano/}}. For comparison, we consider four competitive methods, namely Support Vector regression with Linear kernel (SVLN), Support Vector regression with Radial Basis kernel (SVSIG),  Support Vector regression with Polynomial kernel (SVPOLY), and Multiple Kernel Regression (MKR) \cite{DBLP:conf/kdd/SahooHL14}. 

The parameter selection criterion for the SV-family is that we carefully tune the parameters $\epsilon$ (error margin), $\mathtt{d}$ (degree of kernel function), and $\gamma$ (kernel coefficient) for kernels. Each parameter's value is selected from the sets $\epsilon\in\{10^{-5},10^{-4},...1,...,10^{4},10^{5}\}$, $\mathtt{d} \in \{1,2,3\}$, $\gamma \in\{10^{-5},10^{-4},$ $...1,...,10^{4},10^{5}\}$ respectively, so in total there are 363 combinations for each model. For each test run, during training we iterate through every combination of $\epsilon$,$\mathtt{d}$ and $\gamma$'s candidate values, and keep the values that generate the highest accuracy on the validation set, then use these parameters on the test set and report the results. For comparable evaluation against MKR, we use an offline implementation where test samples are not used to update the parameters, and the number of support vectors is set to 120 for matching the parameter size of TeNet. The hyperparameter selection of TeNet follows the same procedure. We provide more details in Section \ref{ssec:ps}. 

For most of the experiments $d$ is set to 28, meaning for each day, the time-series up to 2pm is known to the model. Selecting this particular number is because considering humans rarely remain active from 12am to 4am and the values in that period are almost all zeros, the first 28d represent information from exactly half of the active period from 4am to 12am of the next day. Such setting is challenging in the sense that the gap between 2pm to 12am next day is substantial and it leaves numerous possible outcomes for the daily accumulation. The complexity involved hence provides insight about how well the proposed and the competitive models can capture an individual's daily patterns and make prediction from limited information. 

Next we present the experimental results for the proposed method and the competitive methods, and also offer in-depth discussion about hyperparameter tuning and about the effect of temporal embedding.

\vspace{-0.2cm}
\subsection{Prediction for Periodical Accumulation}
%\vspace{-0.8cm}
Table \ref{tbl:comp} studies the prediction performances of the proposed method and four competitive methods on 15 individual datasets of three different modalities, evaluated by average HitRate(HR)@20\% and 30\%, Mean Squared Error (MSE) and Mean absolute Relative Error (MRE). Using four metrics is due to that for datasets with long-tailed values (which human behaviors can often be characterized to be \cite{gonzalez2008understanding}), as an absolute measurement, MSE alone is not an ideal metric to evaluate a regression method's performance because it is heavily biased by samples in the long tail \cite{Wang:2013rm,Wang:2014yg}. Therefore we mainly use relative measures for the evaluation while keeping MSE as a reference.

The highlighted numbers in red, black,  magenta and blue indicate the winning performance on that dataset under the corresponding metric ( magenta$\rightarrow$ HR@20\%, blue$\rightarrow$HR@30\%, red$\rightarrow$RE, black$\rightarrow$ MSE). Multiple highlighted numbers with the same color in a row indicate multiple winners under that metric on that dataset. We also report some of the properties, i.e. the total number of samples n, the numeric range [\texttildelow], and standard deviation $\sigma$, for each individual dataset. A closer look at these dataset statistics suggests large varieties in terms of number of samples (from 156 to 874), numerical ranges (0.2 to 345) and variances ( $\sigma$ from 2.8 to 47). To present the reader with more intuitive and meaningful results, the numbers shown are unnormalized. 

\begin{figure}[htp]
\centering
%\hspace{-.3cm}
\vspace{-0.4cm}
    \includegraphics[scale = 0.3]{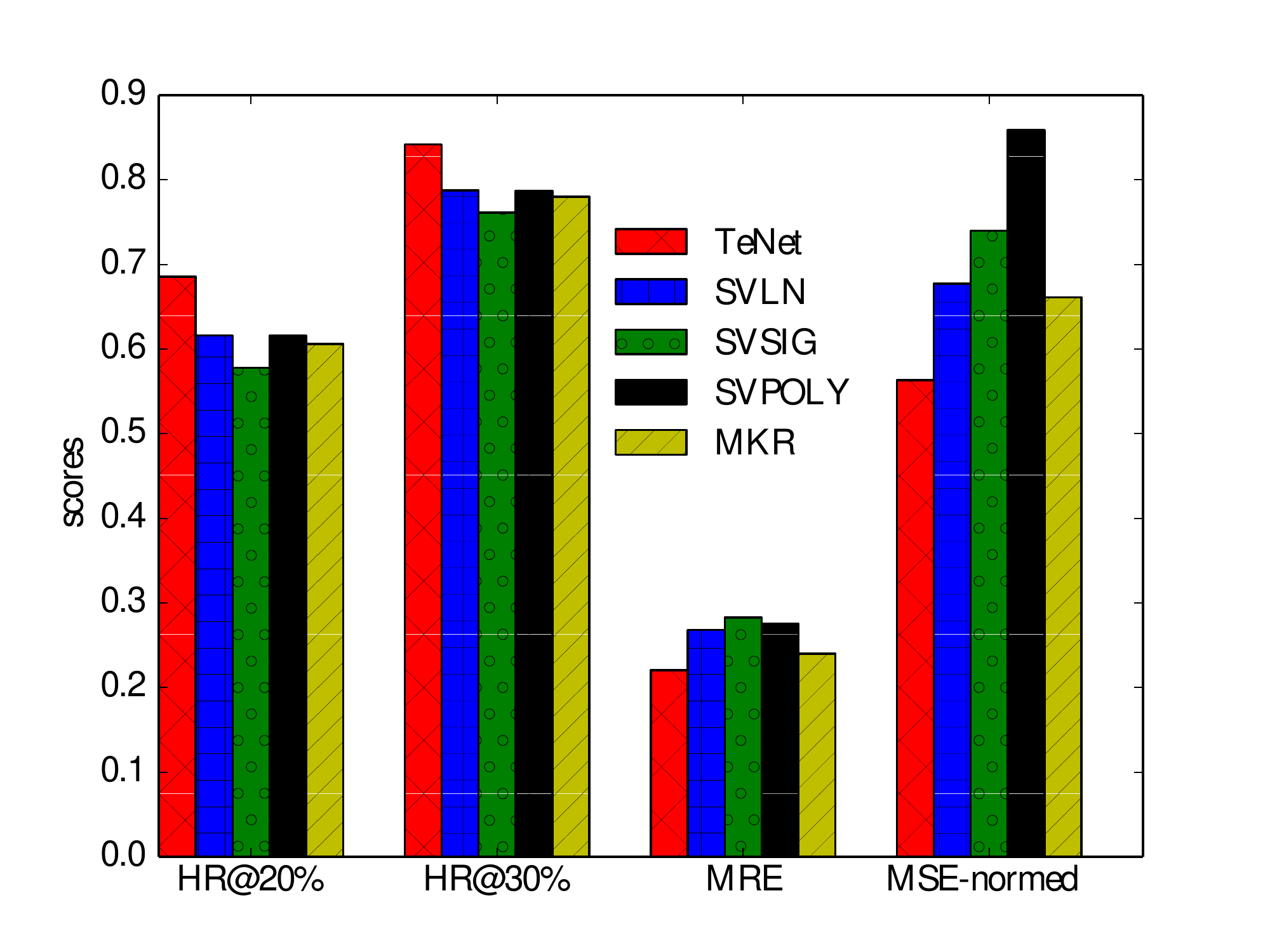}
\vspace{-0.6cm}
\caption{Mean Average Performance}
\label{fig:map}
\end{figure}

\begin{figure}[htp]

%\centering
\vspace{-0.6cm}
\hspace{-.3cm}
    \subfigure[n vs. adv. MRE]{
    \includegraphics[scale = 0.2]{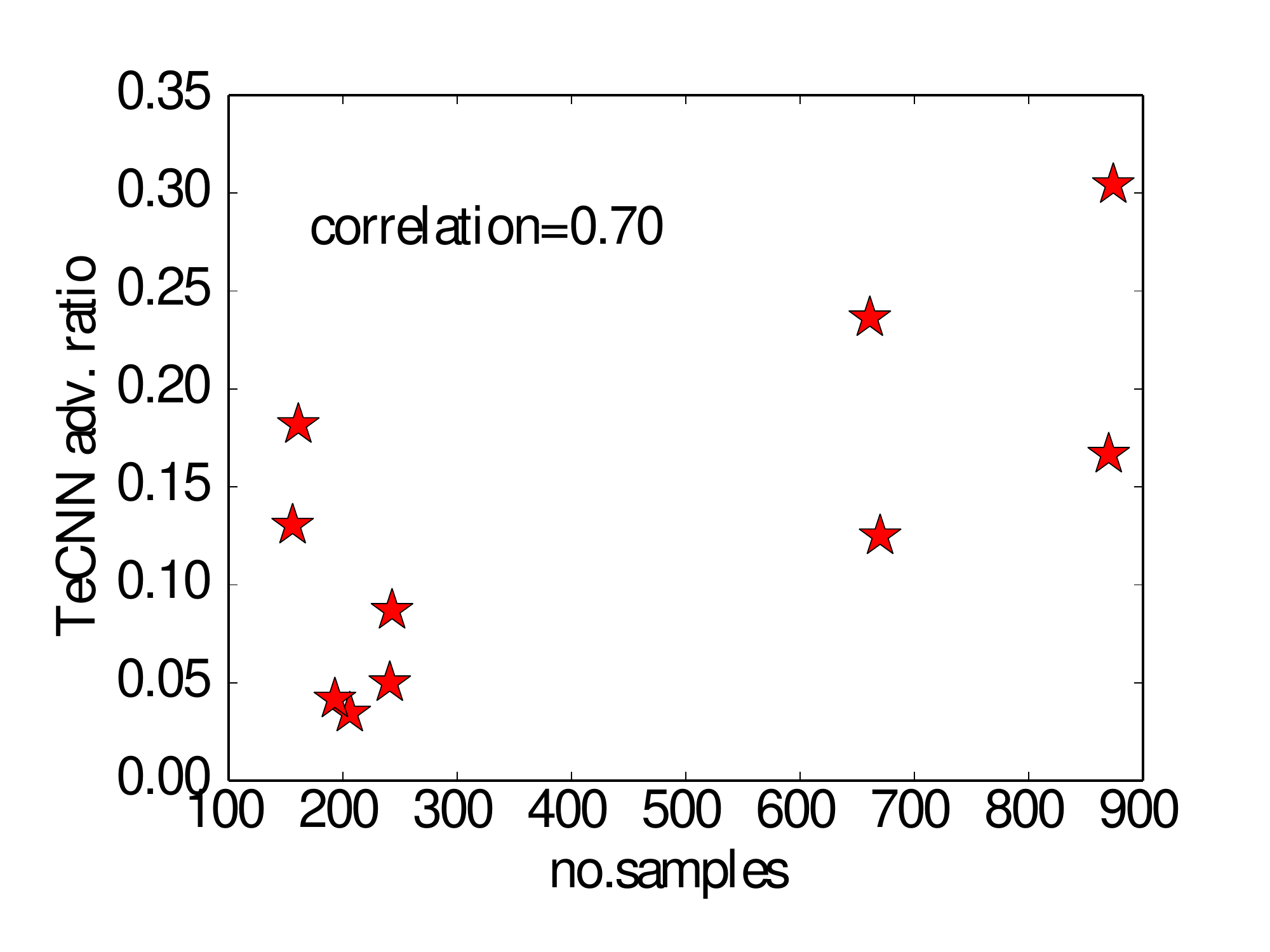} \label{fig:adv1}
    }
    \subfigure[entropy vs. adv. MRE]{
    \includegraphics[scale = 0.2]{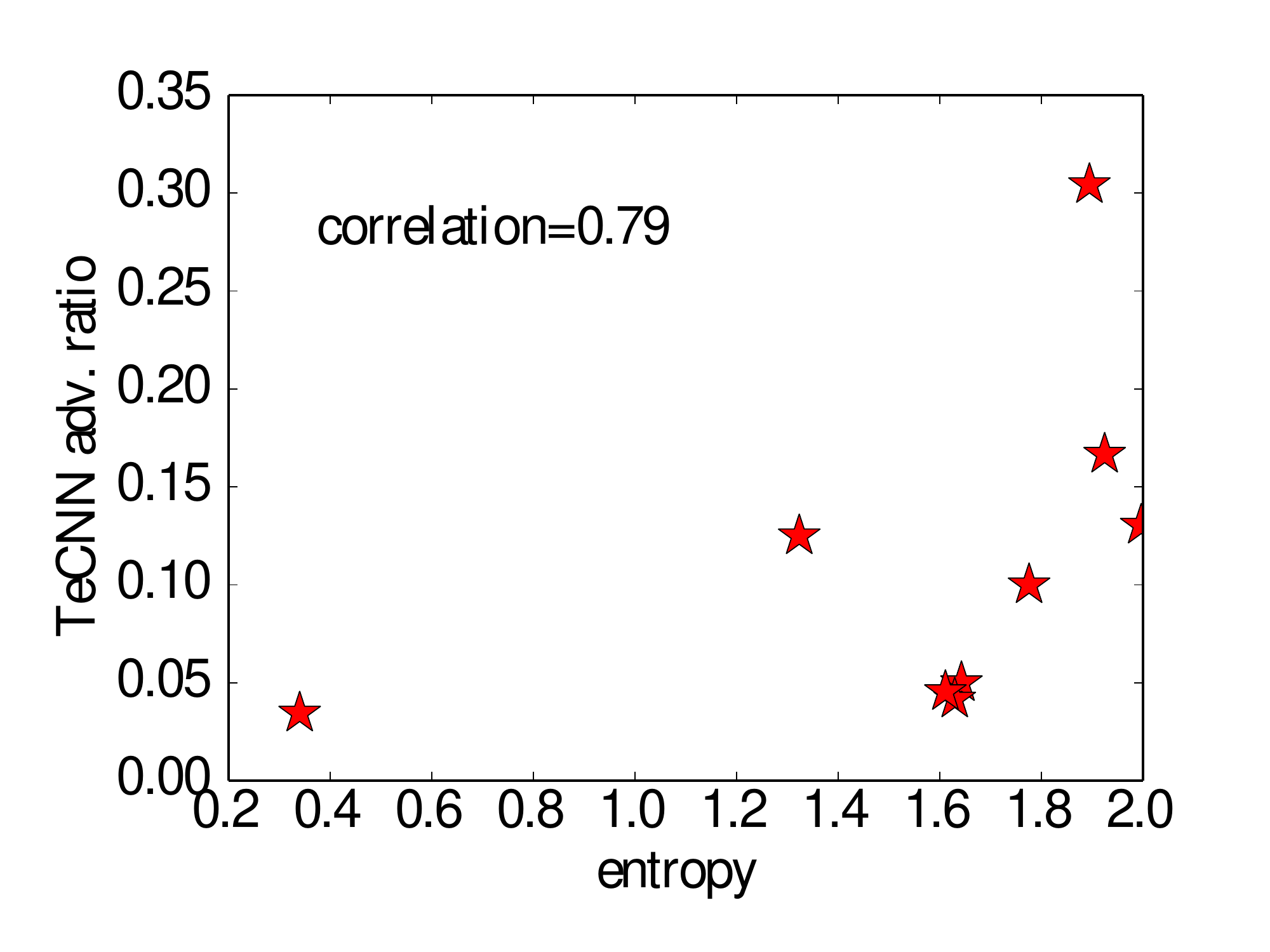} \label{fig:adv2}
    }
\vspace{-0.6cm}
\caption{The correlation between TeNet's performance advantage and sample size/data complexity}
\label{fig:adv}
\end{figure}

\vspace{-0.2cm}
Generally, the distribution of the highlighted and winning performances shows that TeNet achieved best results in most of the cases, with a few but non-systematical exceptions spread across the competitive methods. Out of the 15 individual datasets, TeNet has won 14 entries in HR@20\%, 15 entries in HR@30\%, 13 entries in MRE, and 7 entries in MSE, showing a superior performance among the evaluated models. SVLN and SVSIG show least competitive results by having 1, 0, 2, 1 and  1, 0, 1, 0 winning performances respectively. SVPOLY obtains slightly better results with 3, 2, 3, 0 wins. MKR on the other hand, has shown comparable results in MSE but far less competitive results in other metrics, by having 0, 0, 1, 8 wins. In addition, we find that MKR is less robust to larger numerical ranges such as in HMD-8, HMD-12, HMT-8, and HMT-12, while TeNet demonstrates consistent performances cross all datasets.

To compare the methods quantatitively, we plot Figure \ref{fig:map} and show each method's mean average scores cross all individual datasets (MSE is normalized with the maximum MSE among the methods in each entry). On the 15 individual dataset, TeNet achieved best average performance under all four metrics. Taking a TeNet vs. all approach, we find TeNet's performance and the average of other methods' performance under HR@20\%, HR@30\%, MRE and MSE are 69 vs. 60, 84 vs. 78,  0.22 vs. 0.27 and 34 vs. 51 respectively, showing that TeNet makes a relative improvement of 15\%, 8\%, 19\% and 33\% respectively under the corresponding metric. Then if we investigate TeNet vs. the best among the rest, with HR@20\% 69 and HR@30\% 84, TeNet beats the second best HR@20\% 61 (SVLN, SVPOLY) by 8, the second best HR@30\% 78 (SVLN, SVPOLY) by 6; on MRE and MSE, TeNet's average errors are 0.22 and 34, while the second bests are 0.24 and 40 (MKR). Hence for all 15 individual dataset, in average TeNet marks an 13\% increase in HR@20\%,  an 8\% increase in HR@30\%, a 9.1\% decrease in MRE and a 15\% decrease in MRE to the second best method under each corresponding metric. We also observe that though in all 15 individual datasets TeNet obtained the best performance under HR@30\%, the average winning margin is the smallest than those under other metrics. This is because HR@30\% is a relative looser measurement than other metrics, which leads to the result that less accurate prediction tends to have similar performances. However, the consistent advantage of TeNet in not only HR@30\% but all four metrics still suggests that it has the best prediction accuracy. We hence conclude that TeNet has shown consistent advantages which are robust to variations in the data modality as well as the statistics characteristics of the data.

\begin{figure}[htp]
%\centering
\vspace{-0.3cm}
\hspace{-0.8cm}
    \subfigure[$d$ vs. error]{
    \includegraphics[scale = 0.21]{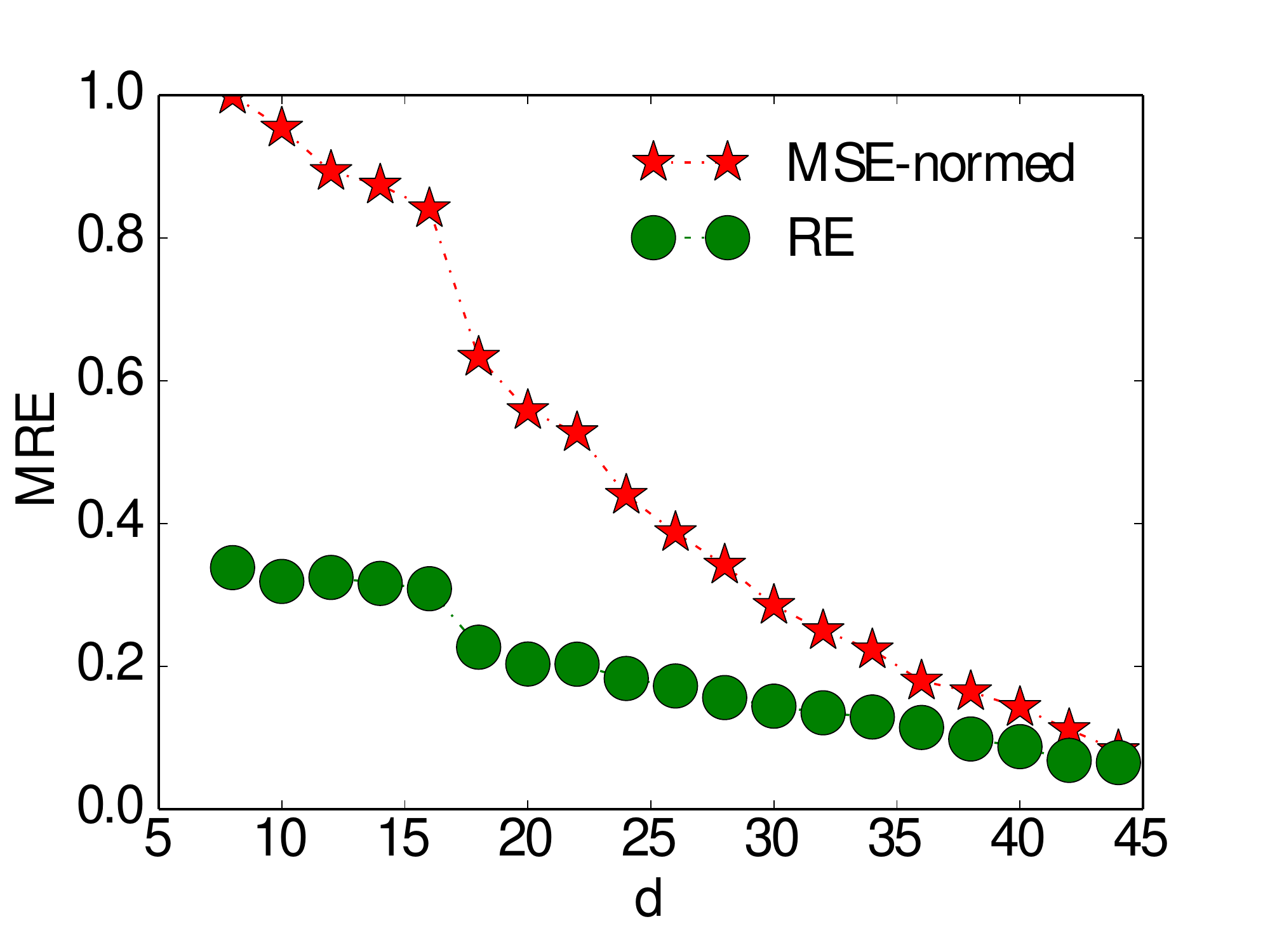}        \label{fig:d2}
    }
        \subfigure[$d$ vs. HR]{
    \includegraphics[scale = 0.21]{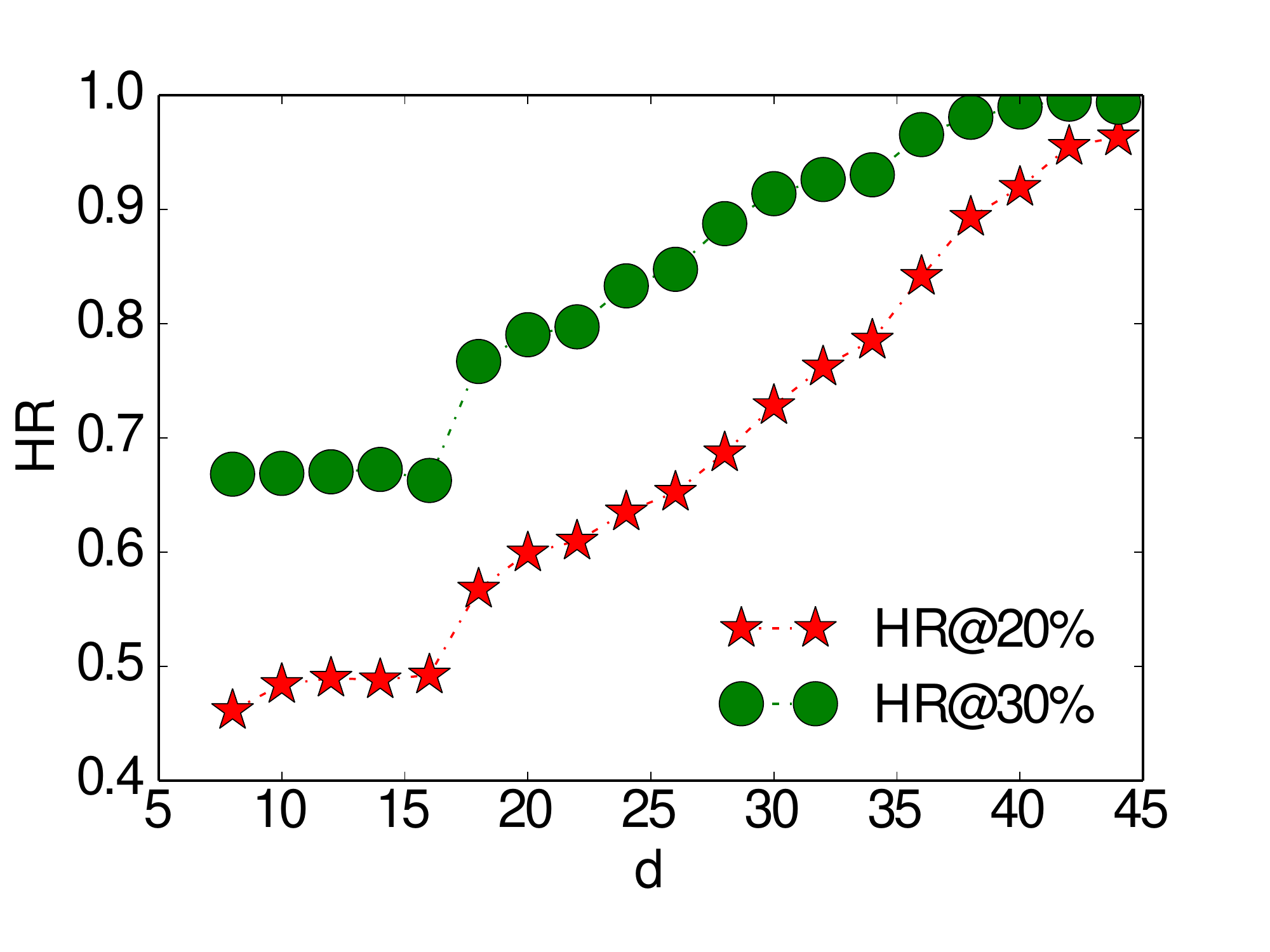}        \label{fig:d3}
    }
    \vspace{-0.5cm}
\caption{The effect of the kernel function degree $d$}
\label{fig:dd}
\end{figure}

\vspace{-0.2cm}
We further examine TeNet's ability to scale up its learning effectiveness with a growing sample size or an increasing complexity of the data. Taking MRE for example,  we measure two correlations using Pearson's correlation coefficient: 1) the correlation between the averaged performance advantage ($\frac{1}{n}\sum_i^n min(\{re_i^{SV*,MKR}\}) - re_i^{TeNet}$) and the sample size, 2) the correlation between the averaged performance advantage and the entropy, for each individual dataset. The measurements yield correlation coefficients 0.7 and 0.79 respectively,  suggesting a strong correlation between each set of the variables.  Such patterns mean that as the sample size or the complexity of the data grows, TeNet is able to learn more effective than other methods to achieve better performance. The correlations are also visually identifiable as we plot the the performance advantage ratios in Figure \ref{fig:adv}.

\vspace{-0.1cm}
\subsection{The Effect of $d$}
Figure \ref{fig:dd} illustrates the effect of the feature dimensionality $d$ on the prediction accuracy. Here we use HPC-AU-8 as a case study. Figure \ref{fig:d2} shows the changes of MRE and normalized MSE to a growing $d$. Unsurprisingly, both errors decrease monotonically as $d$ increases, from 1, 0.35 at $d=8$ to 0.08, 0.07 at $d=44$. Figure \ref{fig:d3} depicts how the HR responds to a growing $d$. Again, we see monotonic growths (almost, except for $d=16$) in HR@20\% and HR@30\%. These results confirm that TeNet can effectively use the additional information and in the mean time has received little impact from the noise in the additional dimensions.
%\subsection{$k$-interval Forecast}
\begin{figure}[htp]
\vspace{-0.3cm}
\centering
        \subfigure[Random Snippets]{
   %     \hspace{-0.5cm}
    \includegraphics[scale = 0.19]{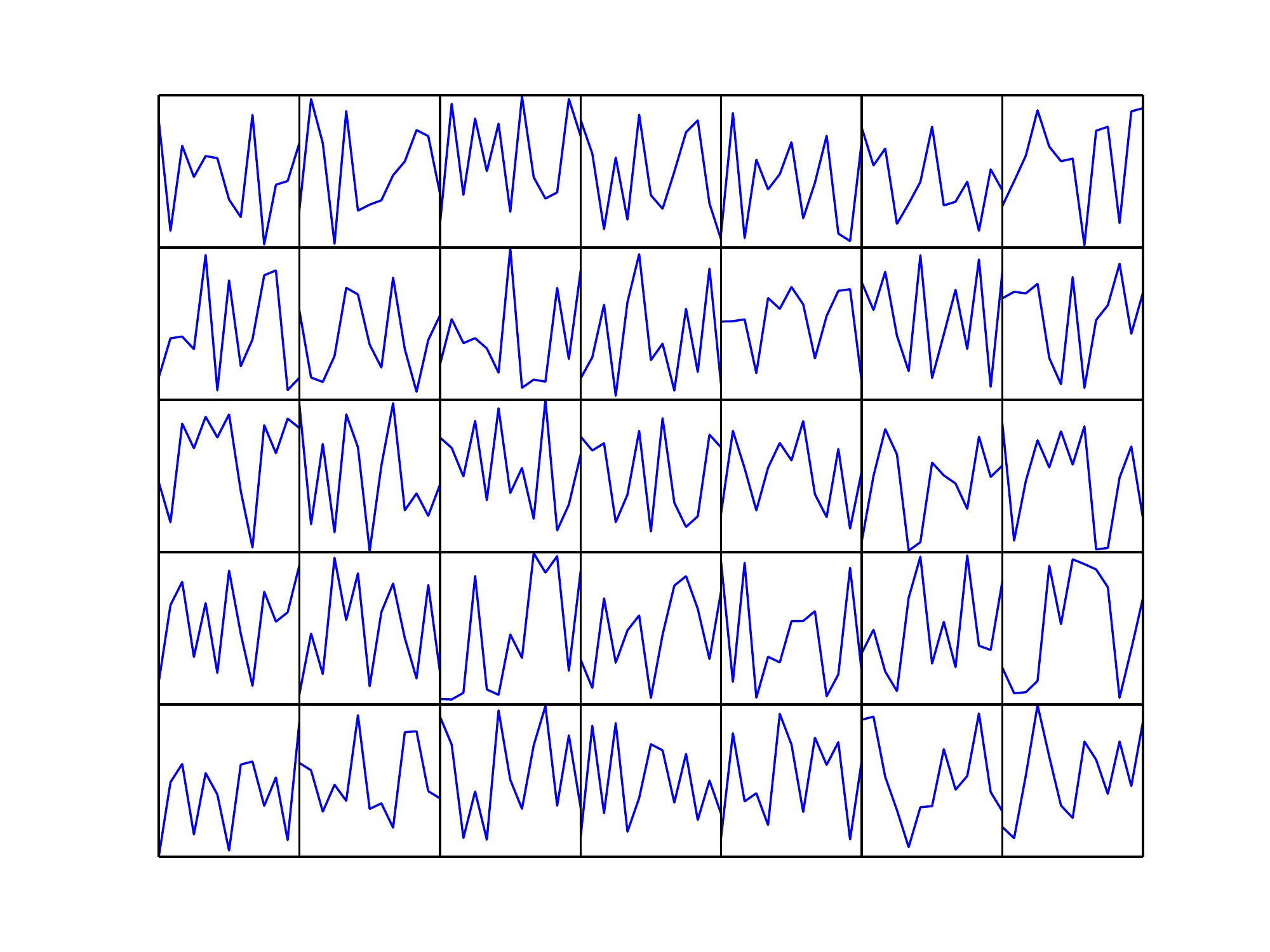}        \label{fig:vs1}
}
    \subfigure[Learned Snippets]{
%    \hspace{-0.5cm}
    \includegraphics[scale = 0.19]{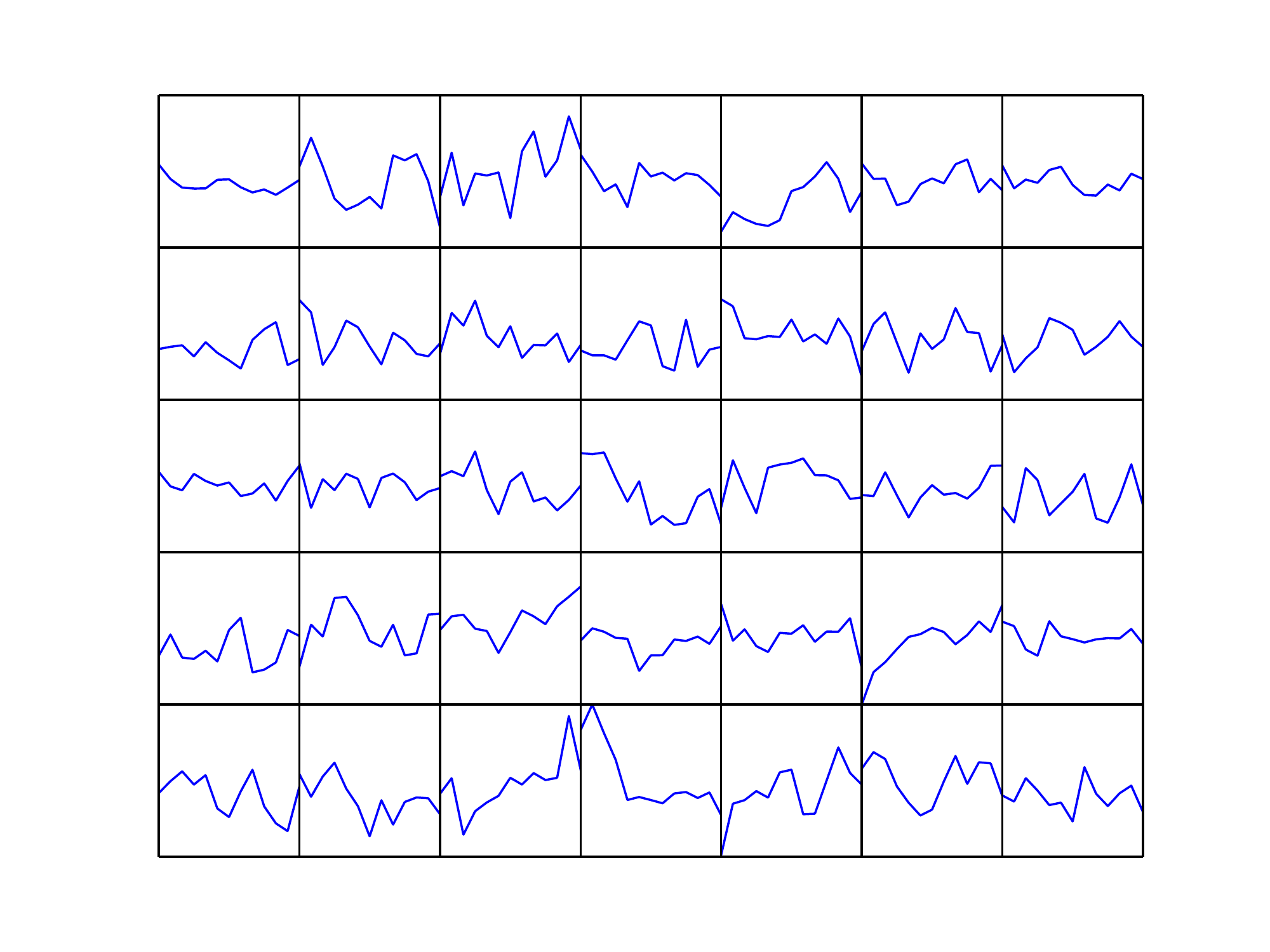}        \label{fig:vs2}
    }
    \vspace{-0.4cm}
\caption{Visualization of a set of learned snippets by TeNet (correspond to the convolution filters)}
\label{fig:vs}
\vspace{-0.6cm}
\end{figure}

%\vspace{-0.6cm}
\subsection{The Effect of Temporal Embedding}
In Section \ref{ssec:te} we discussed how hypothetically temporal embedding would boost the performance of the model by automatically realigning the distorted time-series to the dominating patterns in a dataset, and verified it with a case study on a synthetic example. To further validate this hypothesis on real data, we create a designated dataset from HPC-AU-8 by performing the following procedure:
\begin{enumerate}

\item We run a clustering with the affinity propagation method in \cite{Frey:2007fk}, and find the top 10 exemplars.
\item We take the exemplars and generate 300 synthetic samples (30 for each exemplar) by distorting the exemplars with randomly selected operations such as swapping two neighboring segments or shifting the data forward and backward. They are equally split into training, validation and test set.
\item We train a model with a modified classical convolutional neural network fore regression (CNN, input $\rightarrow$ convolution/pooling $\rightarrow$  sigmoid  $\rightarrow$ l1-linear regression) without temporal embedding, and a model with TeNet, and examine the performance differences.
\end{enumerate}
The results are reported in Table \ref{tbl:tecomp}. We observe that  with the temporal embedding layer, the prediction accuracy has been improved by more than a half (15.5 to 6.4, 0.34 to 0.12) for MSE and MRE, and for about 100\%/40\% in HitRate@20\% and 30\%. This shows that temporal embedding is able to learn the weights which are conceptually equivalent to a reverse operation for the distortions and misalignments.
\begin{table}[htp]
\vspace{-0.9cm}
\centering
\caption{Performance w/\&w/o temporal embedding}
\label{tbl:tecomp}
\small{
\begin{tabular}{ | c |c|c|c|c| }
\hline  &HR@20\% & HR@30\% & MRE & MSE  \\
\hline CNN & 38&66& 0.34 &15.5\\
\hline TeNet & 75& 93& 0.12& 6.4\\
\hline
\end{tabular}
}
\end{table}

\vspace{-0.4cm}
\subsection{Discussion}
\vspace{-0.3cm}
\subsubsection{Distinctive Snippets}
We present a visualization of the random snippets and learned snippets for the first cross-validation iteration on HPC-AU-8 in Figure \ref{fig:vs}. Each cell is a snippet, a segment of time-series the model deems representative. The figures show some noteworthy properties. Firstly the random snippets are fairly dense, while the learned ones are much more sparse, meaning that in most of cases there are only a smaller number of spikes and valleys in each learned snippets. Secondly, the sparsity of the learned snippets is also accompanied by a visually identifiable high distinctiveness across the learned snippets, which means snippets learned tend to be different from one another because they effectively capture different patterns in the training data. Both properties suggest that the snippets are truly learning from the patterns in the dataset and both properties have a positive effect on the model's prediction accuracy.

\vspace{-0.1cm}
\subsubsection{Selection of Hyperparameters}
%\vspace{-0.9cm}
As an issue often posed to complex learning models including neural networks, how to select the hyperparameters is an open question studied by many \cite{DBLP:journals/jmlr/LarochelleBLL09}. There are six hyperparameters in the proposed model:

\begin{table}[htp]
\vspace{-0.6cm}
\caption{Hyperparameters and selection candidates}
\small{
\label{tbl:thp}
\begin{tabular}{ | c |c|l| }
\hline Notation & Description & Candidates \\
\hline $d^f$ & filter size & \{3,5,7\}\\
\hline $n^f$ & no.kernels in conv. layer & \{20,30,40,60\}\\
\hline $r^l$ & learning rate & \{0.01,0.02\}\\
\hline $d^{te}$ & temporal embedding step & \{1,2\}\\
\hline $n^{(3)}$ & no.output in sigmoid layer & \{12,16\}\\
\hline $\lambda$ & weight for the l1-term & \{0.1,0.01,0.001\} \\
\hline
\end{tabular}
}
\end{table}
\label{ssec:ps}
In this paper, since the sizes of the datasets are moderate, we use an intuitive approach to find 
the hyperparameters for the testing. The selection and testing processes follows that described in the third paragraph of Section \ref{ssec:es}. One can also use the greedy hyperparameter selection processed described in  \cite{DBLP:journals/jmlr/LarochelleBLL09}. We also used two optional data preprocessing, i.e. high pass filtering to denoise, and data shifting to synthesize more training data. The activation of each technique is subject to a control parameter which is tuned using the same process. 

Note that since all the hidden nodes in layers 2, 3 output small values only, with the settings we used for experiments, the regression layer's ability to predict larger numbers (e.g. >1000) is limited. To predict larger numbers, one can consider either rescaling the data or setting smaller $\lambda$ to adjust to the numerical range of the specific dataset.

\subsubsection{Network Depth and Number of Parameters}
The proposed model has a moderate number of layers (four if we count the convolution/pooling as one), and hence a moderate number of parameters to estimate. For example, with $d=28$, $d^{te}=1$ (one $W_l$ and one $W_r$),  $n^f$ and $d^f$ set to 20 and 5,  $n^{(3)}=12$, we have:
\begin{eqnarray}
\nonumber  \sum\{|W^*|,|b^*|\} &=& (3\times 28 + 28) + (20\times 5 + 20) \\
\nonumber &&+ (240\times12+12) + (12+1)\\
 &=& 3197
\end{eqnarray}
It is possible to add more layers to construct a deeper architecture based on temporal embedding and convolution. However, the data itself must be complex enough to provide more potential for the model to exploit. Given the granularity of daily human behaviors, for the task of predicting modalities such as traveling distance/time and power consumption, a deeper architecture has only limited effect.

\vspace{-0.1cm}
\section{Conclusion}
\label{sec:conc}
Motivated by the observation that regularities in periodical time-series sometimes manifest at different moments and at varied paces, in this paper we propose a technique called temporal embedding and devise a convolutional neural network-based learning model called TeNet, which is robust to temporal distortions and misalignments,  to learn abstract features.  First we  present TeNet and discuss the intuition behind it using a case study, and then describe the technical details for the whole network architecture, and solve the backpropagation problem for the proposed model. In the experiments we use an extensive range of real-life periodical data that covers three modalities to compare the performances of the proposed model against competitive methods. We find that in  average TeNet achieves 8\% to 33\% advantage against other methods in difference metrics and the advantage scales up with a growing sample size used in training. We also find that the accuracy of TeNet increases almost monotonically with a growing $d$, indicating the model is effective in utilizing more information and while remaining robust to noise. We also create a set of synthetic data from the real-life data to demonstrate the effect of temporal embedding and successfully show its capability of realigning distorted and misaligned data. At the end of the experiment we also offer an in-depth discussion about hyperparameter selection, data preprocessing, network depth and number of parameters, and present a visualization of the learned snippets. Beyond the periodical accumulation prediction problem, we expect Tenet to be useful for general time-series predictions ranging from forecasts to k-ahead prediction. 

\vspace{-0.2cm}
\small{
\bibliographystyle{abbrv}
\bibliography{kdd15}

\begin{thebibliography}{10}

\bibitem{DBLP:journals/taslp/Abdel-HamidMJDPY14}
O.~Abdel{-}Hamid, A.~Mohamed, H.~Jiang, L.~Deng, G.~Penn, and D.~Yu.
\newblock Convolutional neural networks for speech recognition.
\newblock {\em {IEEE/ACM} Transactions on Audio, Speech {\&} Language
  Processing}, 22(10):1533--1545, 2014.

\bibitem{DBLP:journals/ftml/Bengio09}
Y.~Bengio.
\newblock Learning deep architectures for {AI}.
\newblock {\em Foundations and Trends in Machine Learning}, 2(1):1--127, 2009.

\bibitem{bengio2007greedy}
Y.~Bengio, P.~Lamblin, D.~Popovici, and H.~Larochelle.
\newblock Greedy layer-wise training of deep networks.
\newblock {\em NIPS}, 19:153, 2007.

\bibitem{yonsei-lifemap-2012-01-03}
Y.~Chon, E.~Talipov, H.~Shin, and H.~Cha.
\newblock {CRAWDAD} data set yonsei/lifemap (v. 2012-01-03).
\newblock Downloaded from http://crawdad.org/yonsei/lifemap/, Jan. 2012.

\bibitem{DBLP:conf/ijcai/CiresanMMGS11}
D.~C. Ciresan, U.~Meier, J.~Masci, L.~M. Gambardella, and J.~Schmidhuber.
\newblock Flexible, high performance convolutional neural networks for image
  classification.
\newblock In {\em IJCAI}, pages 1237--1242, 2011.

\bibitem{collobert2008unified}
R.~Collobert and J.~Weston.
\newblock A unified architecture for natural language processing: Deep neural
  networks with multitask learning.
\newblock In {\em ICML}, pages 160--167. ACM, 2008.

\bibitem{DBLP:journals/jmlr/CollobertWBKKK11}
R.~Collobert, J.~Weston, L.~Bottou, M.~Karlen, K.~Kavukcuoglu, and P.~P. Kuksa.
\newblock Natural language processing (almost) from scratch.
\newblock {\em Journal of Machine Learning Research}, 12:2493--2537, 2011.

\bibitem{DBLP:conf/icassp/DengLHYYSSZHWGA13}
L.~Deng, J.~Li, J.~Huang, K.~Yao, D.~Yu, F.~Seide, M.~L. Seltzer, G.~Zweig,
  X.~He, J.~Williams, Y.~Gong, and A.~Acero.
\newblock Recent advances in deep learning for speech research at microsoft.
\newblock In {\em ICASSP}, pages 8604--8608, 2013.

\bibitem{DBLP:journals/pr/FischerI14}
A.~Fischer and C.~Igel.
\newblock Training restricted boltzmann machines: An introduction.
\newblock {\em Pattern Recognition}, 47(1):25--39, 2014.

\bibitem{Frey:2007fk}
B.~J. Frey and D.~Dueck.
\newblock Clustering by passing messages between data points.
\newblock {\em Science}, 315(5814):972--976, February 2007.

\bibitem{gonzalez2008understanding}
M.~C. Gonzalez, C.~A. Hidalgo, and A.-L. Barabasi.
\newblock Understanding individual human mobility patterns.
\newblock {\em Nature}, 453(7196):779--782, 2008.

\bibitem{DBLP:journals/corr/HannunCCCDEPSSCN14}
A.~Y. Hannun, C.~Case, J.~Casper, B.~C. Catanzaro, G.~Diamos, E.~Elsen,
  R.~Prenger, S.~Satheesh, S.~Sengupta, A.~Coates, and A.~Y. Ng.
\newblock Deep speech: Scaling up end-to-end speech recognition.
\newblock {\em CoRR}, abs/1412.5567, 2014.

\bibitem{DBLP:series/lncs/Hinton12}
G.~E. Hinton.
\newblock A practical guide to training restricted boltzmann machines.
\newblock In G.~Montavon, G.~B. Orr, and K.~M{\"{u}}ller, editors, {\em Neural
  Networks: Tricks of the Trade - Second Edition}, volume 7700 of {\em Lecture
  Notes in Computer Science}, pages 599--619. Springer, 2012.

\bibitem{Jurdak13}
R.~Jurdak, P.~Sommer, B.~Kusy, N.~Kottege, C.~Crossman, A.~Mckeown, and
  D.~Westcott.
\newblock Camazotz: Multimodal activity-based gps sampling.
\newblock In {\em IPSN}, 2013.

\bibitem{DBLP:journals/jmlr/LarochelleBLL09}
H.~Larochelle, Y.~Bengio, J.~Louradour, and P.~Lamblin.
\newblock Exploring strategies for training deep neural networks.
\newblock {\em Journal of Machine Learning Research}, 10:1--40, 2009.

\bibitem{726791}
Y.~Lecun, L.~Bottou, Y.~Bengio, and P.~Haffner.
\newblock Gradient-based learning applied to document recognition.
\newblock {\em Proceedings of the IEEE}, 86(11):2278--2324, Nov 1998.

\bibitem{DBLP:conf/nips/LeeEN07}
H.~Lee, C.~Ekanadham, and A.~Y. Ng.
\newblock Sparse deep belief net model for visual area {V2}.
\newblock In {\em NPIS}, pages 873--880, 2007.

\bibitem{lee2009convolutional}
H.~Lee, R.~Grosse, R.~Ranganath, and A.~Y. Ng.
\newblock Convolutional deep belief networks for scalable unsupervised learning
  of hierarchical representations.
\newblock In {\em ICML}, pages 609--616, 2009.

\bibitem{lee2009unsupervised}
H.~Lee, P.~Pham, Y.~Largman, and A.~Y. Ng.
\newblock Unsupervised feature learning for audio classification using
  convolutional deep belief networks.
\newblock In {\em NIPS}, pages 1096--1104, 2009.

\bibitem{DBLP:journals/nn/MatsuguMMK03}
M.~Matsugu, K.~Mori, Y.~Mitari, and Y.~Kaneda.
\newblock Subject independent facial expression recognition with robust face
  detection using a convolutional neural network.
\newblock {\em Neural Networks}, 16(5-6):555--559, 2003.

\bibitem{ng2004feature}
A.~Y. Ng.
\newblock Feature selection, l 1 vs. l 2 regularization, and rotational
  invariance.
\newblock In {\em ICML}, page~78, 2004.

\bibitem{DBLP:conf/icml/NgiamKKNLN11}
J.~Ngiam, A.~Khosla, M.~Kim, J.~Nam, H.~Lee, and A.~Y. Ng.
\newblock Multimodal deep learning.
\newblock In {\em ICML}, pages 689--696, 2011.

\bibitem{DBLP:conf/kdd/RakthanmanonCMBWZZK12}
T.~Rakthanmanon, B.~J.~L. Campana, A.~Mueen, G.~E. A. P.~A. Batista, M.~B.
  Westover, Q.~Zhu, J.~Zakaria, and E.~J. Keogh.
\newblock Searching and mining trillions of time series subsequences under
  dynamic time warping.
\newblock In {\em SIGKDD}, pages 262--270, 2012.

\bibitem{DBLP:conf/kdd/SahooHL14}
D.~Sahoo, S.~C.~H. Hoi, and B.~Li.
\newblock Online multiple kernel regression.
\newblock In {\em SIGKDD}, pages 293--302, 2014.

\bibitem{schmidt2005least}
M.~Schmidt.
\newblock Least squares optimization with l1-norm regularization.
\newblock {\em CS542B Project Report}, 2005.

\bibitem{schneider2013unravelling}
C.~M. Schneider, V.~Belik, T.~Couronn{\'e}, Z.~Smoreda, and M.~C. Gonz{\'a}lez.
\newblock Unravelling daily human mobility motifs.
\newblock {\em Journal of The Royal Society Interface}, 10(84):20130246, 2013.

\bibitem{schneider2013daily}
C.~M. Schneider, C.~Rudloff, D.~Bauer, and M.~C. Gonz{\'a}lez.
\newblock Daily travel behavior: Lessons from a week-long survey for the
  extraction of human mobility motifs related information.
\newblock In {\em ACM SIGKDD International Workshop on Urban Computing},
  page~3, 2013.

\bibitem{DBLP:conf/acl/TsuruokaTA09}
Y.~Tsuruoka, J.~Tsujii, and S.~Ananiadou.
\newblock Stochastic gradient descent training for l1-regularized log-linear
  models with cumulative penalty.
\newblock In {\em {ACL} 2009}, pages 477--485, 2009.

\bibitem{DBLP:conf/icml/VincentLBM08}
P.~Vincent, H.~Larochelle, Y.~Bengio, and P.~Manzagol.
\newblock Extracting and composing robust features with denoising autoencoders.
\newblock In {\em ICML}, pages 1096--1103, 2008.

\bibitem{DBLP:journals/jmlr/VincentLLBM10}
P.~Vincent, H.~Larochelle, I.~Lajoie, Y.~Bengio, and P.~Manzagol.
\newblock Stacked denoising autoencoders: Learning useful representations in a
  deep network with a local denoising criterion.
\newblock {\em Journal of Machine Learning Research}, 11:3371--3408, 2010.

\bibitem{Wang:2013rm}
D.~Wang, C.~Song, and A.-L. Barabasi.
\newblock Quantifying long-term scientific impact.
\newblock {\em Science}, 342(6154):127--132, October 2013.

\bibitem{Wang:2014yg}
D.~Wang, C.~Song, H.-W. Shen, and A.-L. Barabasi.
\newblock Response to comment on "quantifying long-term scientific impact".
\newblock {\em Science}, 345(6193), July 2014.

\bibitem{zandbergen2009accuracy}
P.~A. Zandbergen.
\newblock Accuracy of iphone locations: A comparison of assisted gps, wifi and
  cellular positioning.
\newblock {\em Transactions in GIS}, 13(s1):5--25, 2009.

\bibitem{jrsi}
K.~Zhao, R.~Jurdak, J.~Liu, D.~Westcott, B.~Kusy, H.~Parry, P.~Sommer, and
  A.~McKeown.
\newblock Optimal l{\'e}vy-flight foraging in a finite landscape.
\newblock {\em Journal of The Royal Society Interface}, 12(104), 2015.

\end{thebibliography}
}
\end{document}